Multi-modal, multi-task, multi-attention (M3) deep learning detection of reticular pseudodrusen: towards automated and accessible classification of age-related macular degeneration

Qingyu Chen, PhD[1]*, Tiarnan D. L. Keenan, BM BCh, PhD[2]*, Alexis Allot, PhD[1], Yifan Peng, PhD[1], Elvira Agrón, MA[2], Amitha Domalpally, MD, PhD[3], Caroline C. W. Klaver, MD, PhD[4], Daniel T. Luttikhuizen[4], Marcus H. Colyer, MD[5], Catherine A. Cukras, MD, PhD[2], Henry E. Wiley, MD[2], M. Teresa Magone, MD[2], Chantal Cousineau-Krieger, MD[2], Wai T. Wong MD, PhD[2,6], Yingying Zhu[7,8], PhD Emily Y. Chew, MD[2†], Zhiyong Lu, PhD[1†], for the AREDS2 Deep Learning Research Group[9]

1. National Center for Biotechnology Information, National Library of Medicine, National Institutes of Health (NIH), Bethesda, MD, USA

2. Division of Epidemiology and Clinical Applications, National Eye Institute, National Institutes of Health, Bethesda, MD, USA

3. Fundus Photograph Reading Center, University of Wisconsin, Madison, WI, USA

4. Department of Ophthalmology, Erasmus Medical Center, Rotterdam, Netherlands

5. Department of Surgery, Uniformed Services University of the Health Sciences, Bethesda, MD, USA

6. Section on Neuron-Glia Interactions in Retinal Disease, Laboratory of Retinal Cell and Molecular Biology, National Eye Institute, National Institutes of Health, Bethesda, MD, USA

7. Department of Computer Science and Engineering, University of Texas at Arlington, Arlington, TX, USA

8. Department of Radiology, Clinical Center, National Institutes of Health, Bethesda, MD, USA

9. See appendix

* These authors contributed equally to this work

† To whom correspondence should be addressed: echew@nei.nih.gov; zhiyong.lu@nih.gov

**Abstract**

*Objective*

Reticular pseudodrusen (RPD), a key feature of age-related macular degeneration (AMD), are poorly detected by human experts on standard color fundus photography (CFP) and typically require advanced imaging modalities such as fundus autofluorescence (FAF). The objective was to develop and evaluate the performance of a novel 'M3' deep learning framework on RPD detection.

*Materials and Methods*

A deep learning framework M3 was developed to detect RPD presence accurately using CFP alone, FAF alone, or both, employing >8000 CFP-FAF image pairs obtained prospectively (Age-Related Eye Disease Study 2). The M3 framework includes multi-modal (detection from single or multiple image modalities), multi-task (training different tasks simultaneously to improve generalizability), and multi-attention (improving ensembled feature representation) operation. Performance on RPD detection was compared with state-of-the-art deep learning models and 13 ophthalmologists; performance on detection of two other AMD features (geographic atrophy and pigmentary abnormalities) was also evaluated.

*Results*

For RPD detection, M3 achieved area under receiver operating characteristic (AUROC) 0.832, 0.931, and 0.933 for CFP alone, FAF alone, and both, respectively. M3 performance on CFP was very substantially superior to human retinal specialists (median F1-score 0.644 versus 0.350). External validation (on Rotterdam Study, Netherlands) demonstrated high accuracy on CFP alone (AUROC 0.965). The M3 framework also accurately detected geographic atrophy and pigmentary abnormalities (AUROC 0.909 and 0.912, respectively), demonstrating its generalizability.

*Conclusion*

This study demonstrates the successful development, robust evaluation, and external validation of a novel deep learning framework that enables accessible, accurate, and automated AMD diagnosis and prognosis.

**INTRODUCTION**

Age-related macular degeneration (AMD) is the leading cause of legal blindness in developed countries [1 2]. Late AMD is the stage with the potential for severe visual loss; it takes two forms, geographic atrophy and neovascular AMD. AMD is traditionally diagnosed and classified using color fundus photography (CFP) [3], the most widely used and accessible imaging modality in ophthalmology. In the absence of late disease, two main features (macular drusen and pigmentary abnormalities) are used to classify disease and stratify risk of progression to late AMD [3]. More recently, additional imaging modalities have become available in specialist centers, particularly fundus autofluorescence (FAF) imaging [4 5]. Following these developments in retinal imaging, a third macular feature (reticular pseudodrusen, RPD) is now recognized as a key AMD lesion [6 7]. RPD presence is strongly and independently associated with increased risk of progression to late AMD, including two-fold increased risk of geographic atrophy [6], as well as faster enlargement of geographic atrophy [8], which is an important endpoint in ongoing clinical trials. However, RPD are very poorly visible to human eyes on clinical examination or CFP, even to trained experts at the reading center level [9-11].

CFP and FAF are considered complementary imaging modalities [12]. In AMD, some disease features are visualized more clearly to human experts on one or the other modality. For example, macular drusen are typically observed well on CFP but poorly on FAF, while the opposite is true for RPD [9-12]. Other AMD features are observed on both modalities. For example, pigmentary abnormalities are seen on both (though typically classified on CFP) [3 12], while geographic atrophy is seen on both (but typically identified and measured on FAF) [12 13]. Importantly, while CFP is easily performed and very widely available across the globe, FAF imaging is usually available only at specialized academic centers in the developed world; even there, FAF imaging lies outside current standards of care. Hence, any techniques that enable the accurate ascertainment of the full spectrum of AMD features (particularly RPD) from CFP alone would be extremely valuable for improved disease classification and risk prediction.

Recent deep learning approaches have been proposed for the diagnosis and classification of AMD, based on CFP, specifically in the automated detection of macular drusen, pigmentary abnormalities, and geographic atrophy [14-18]. However, several problems apply to these approaches. First, we are not aware of deep learning approaches to RPD detection from CFP or FAF images by other groups. Second, these approaches have generally not incorporated CFP and FAF images together, so the phenotypic characterization of disease is partially limited. Third, very few studies have reported the results of external validation, i.e., where models were tested on a distinct population not used for training, so the possibility of model overfitting is high.

For these reasons, we have developed, trained, and tested a new deep learning approach that benefits from multi-modal, multi-task, multi-attention (M3) operation. The multi-modal operation means that the trained models are versatile and can handle three different image scenarios in practical use (i.e., using CFP alone, FAF alone, or both together as input). The aim of the multi-modal and multi-task training was for single image modality models (either CFP-alone or FAF-alone) to benefit from the complementary information present in both image types during training. Essentially, what is learned for each image modality task can assist during training for the other image modality tasks (by sharing features that are generalizable between the image modalities). Hence, even when the approach is used clinically in the CFP-alone scenario, this benefit is retained. In addition, we employed a multi-attention mechanism. It firstly uses self-attention [19] to distill important features for each modality, which makes the framework suitable when only single-modality images are available. Then, it uses cross-modality attention [20] to ensemble the distilled features from different modalities. The multi-attention mechanism improves the interpretability of image features for diagnosis.

To test the potential benefits of this new M3 approach, we compared the performance of the M3 models with existing state-of-the-art deep learning models, for each of the three image scenarios. We further compared the performance of the M3 models with a total of 13 ophthalmologists at three different expertise levels. We then performed external validation of the M3 models by testing on an independent, well-characterized RPD image dataset from a different continent. (A user-friendly desktop application was developed for enabling external evaluation by end users; the software tool is available upon request). To demonstrate that the M3 technique is generalizable to different tasks and datasets, we also applied this method to two other important AMD features: geographic atrophy and pigmentary abnormalities.

*Background*

Most deep learning models in the medical informatics domain are single-modality approaches, as demonstrated in recent literature reviews [21-25]. For instance, a detailed review summarized 15 deep learning models for retinal disease diagnosis; all were single-modality, using either CFP alone or optical coherence tomography (OCT) images alone [22]. By contrast, multi-modal deep learning models, where image features from different modalities are captured and fused, have been used more widely in general medical image applications, such as tumor image segmentation (computed tomography (CT), magnetic resonance (MR), and positron emission tomography (PET) image tuples) [26], lung image retrieval (CT and PET pairs) [27], and Alzheimer's disease diagnosis (PET and MR pairs) [28]. These studies demonstrate that the performance of multi-modal deep learning models is more effective than their counterparts using single-modality images. Indeed, recent reviews in the ophthalmology domain consider developing multi-modal deep learning models as an important future direction [22 24 25].

To our knowledge, few studies have used multi-modal deep learning models for retinal disease diagnosis. We are aware of only two such studies for the detection of AMD: Vaghefi *et al* trained a deep learning model using CFP, OCT, and OCT-angiography image tuples to identify the presence of intermediate AMD [29], and Yoo *et al* trained a deep learning model using CFP and OCT image pairs to detect the presence of AMD [30]. While both studies reported that the multi-modal deep learning models achieved higher performance than single-modality models, they have important limitations. First, these previous models require that all image modalities used during training must be present during operation. For example, if a model was trained using CFP, OCT, and OCT-angiography images, all three image types must be available for classification. This may not be practical, particularly for advanced imaging modalities like OCT-angiography or FAF, so the accessibility of models like this may remain very limited. Second, the existing methods simply concatenated the image features extracted from the different modalities. This is not effective, especially when the modalities are very different [31 32].

In response, we proposed a novel multi-modal deep learning framework, which uses a multi-task learning technique and a multi-attention mechanism for detecting macular features with improved performance. Notably, this work complements our previous work [33], where we fine-tuned convolutional neural network (CNN) models such as InceptionV3 to detect the presence of RPD on single modality images.

**MATERIALS AND METHODS**

*Primary dataset for deep learning model training and internal validation*

The primary dataset used for deep learning model training and internal validation was the dataset of images, labels, and accompanying clinical information from the Age-Related Eye Disease Study 2 (AREDS2). The AREDS2 was a multicenter phase III randomized controlled clinical trial designed to assess the effects of nutritional supplements on the course of AMD in people at moderate to high risk of progression to late AMD [34]. Its primary outcome was the development of late AMD, defined as neovascular AMD or central geographic atrophy. Institutional review board approval was obtained at each clinical site and written informed consent for the research was obtained from all study participants. The research was conducted under the Declaration of Helsinki and complied with the Health Insurance Portability and Accountability Act.

The AREDS2 study design has been described previously [34]. In short, 4,203 participants aged 50 to 85 years were recruited between 2006 and 2008 at 82 retinal specialty clinics in the United States. Inclusion criteria at enrollment were the presence of either bilateral large drusen, or late AMD in one eye and large drusen in the fellow eye. At baseline and annual study visits, comprehensive eye examinations were performed by certified study personnel using standardized protocols. The study visits included the capture of digital CFP by certified technicians using standard imaging protocols. In the current study, the field 2 images (i.e., 30 degree imaging field centered on the fovea) were used.

In addition, as described previously [11], the AREDS2 ancillary study of FAF imaging was conducted at 66 selected clinic sites, according to the availability of imaging equipment. Sites were permitted to join the ancillary study at any time after FAF imaging equipment became available during the five-year study period. The FAF images were acquired from the Heidelberg Retinal Angiograph (Heidelberg Engineering, Heidelberg, Germany) and fundus cameras with autofluorescence capability by certified technicians using standard imaging protocols. For the Heidelberg images, a single image was acquired at 30 degrees centered on the fovea, captured in high speed mode (768 x768 pixels), using the automated real time mean function set at 14. All images (both CFP and FAF) were sent to the University of Wisconsin Fundus Photograph Reading Center.

The primary dataset consisted of all AREDS2 images where a CFP-FAF pair was available, i.e., where a CFP and a corresponding FAF image (taken from the same eye at the same study visit) were available. The dataset is described with these CFP-FAF pairs as the imaging unit. The total number of images was 8487 (i.e. 8487 CFP, 8487 FAF images, and 8487 CFP-FAF image pairs). The dataset was split randomly into three sets, with the division made at the participant level (such that all images from a single participant were present in one of the three sets only): 70% for training, 10% for validation, and 20% for testing of the models. The characteristics of the participants and images used for training and testing are shown in Table 1.

Table 1. Numbers of study participants and color fundus photograph/fundus autofluorescence image pairs used for deep learning model training. The full set of image pairs was divided into the following subsets at the participant level: training set (70% of the participants), validation set (10%), and test set (20%).

|  | Training set | Validation set | Test set | Total |
|---|---|---|---|---|
| **Participants** (n) | 1,541 | 212 | 433 | 2,186 |
| Female sex (%) | 57.7 | 56.1 | 56.6 | 57.3 |
| Mean age (years) | 72.4 | 72.7 | 73.2 | 72.6 |
| **Image pairs** (n)* | 5,966 | 838 | 1,683 | 8,487 |
| Reticular pseudodrusen present (%)† | 28.3 | 25.7 | 27.6 | 27.9 |
| Geographic atrophy present (%)‡ | 18.8 | 18.6 | 21.6 | 19.4 |
| Pigmentary abnormalities present (%)‡ | 82.8 | 80.5 | 83.6 | 82.7 |

* These refer to the total number of color fundus photograph and corresponding fundus autofluorescence image pairs, i.e., 8,487 color fundus photographs and the 8,487 corresponding fundus autofluorescence images that were captured on the same eye at the same study visit.
† According to reading center expert grading of the fundus autofluorescence images, which provided the ground truth labels for presence/absence of reticular pseudodrusen.
‡ According to reading center expert grading of the color fundus photographs, which provided the ground truth labels for presence/absence of both geographic atrophy and pigmentary abnormalities.

*Ground truth labels for the primary dataset*

The ground truth labels used for training and testing were the grades previously assigned to the images by expert human graders at the University of Wisconsin Fundus Photograph Reading Center. The protocol and definitions used for RPD grading have been described recently [11]. In brief, the RPD grading was performed from FAF images (since RPD are detected by human experts with far greater accuracy on FAF images than on CFP [9-11]). The expert human grading team comprised six graders: four primary graders and two senior graders as adjudicators. These graders did not overlap with the 13 ophthalmologists described elsewhere. RPD were defined as clusters of discrete round or oval lesions of hypoautofluorescence, usually similar in size, or confluent ribbon-like patterns with intervening areas of normal or increased autofluorescence; a minimum of 0.5 disc areas (approximately five lesions) was required. Two primary graders at the reading center independently evaluated FAF images for the presence of RPD; in the case of disagreement between the two primary graders, a senior grader at the reading center would adjudicate the final grade. Inter-grader agreement for the presence/absence of RPD was 94% [11]. Label transfer was used between the FAF images and their corresponding CFP images; this means that the ground truth label obtained from the reading center for each FAF image was also applied to the corresponding CFP. Similarly, the labels from the FAF images were also applied to the CFP-FAF image pairs.

The protocol and definitions used for the grading of (i) geographic atrophy and (ii) pigmentary abnormalities have been described previously [35]. This grading was performed from CFP (since this remains the gold standard for grading pigmentary abnormalities and was traditionally considered the gold standard for grading geographic atrophy). For the baseline images, two primary graders independently evaluated the CFP for these two features; in the case of disagreement between the two primary graders, a senior grader at the reading center would adjudicate the final grade. For the annual follow-up images, a single grader performed grading (independent of the images and grading results from previous visits). Geographic atrophy was defined as an area of partial or complete depigmentation of the retinal pigment epithelium, of size equal to or larger than drusen circle I-2 (diameter 433 mm, area 0.146 mm$^2$, i.e. 1/4 disc diameter and 1/16 disc area) at its widest diameter, with at least two of the following features: roughly circular or oval shape, well-demarcated margins, and visibility of underlying large choroidal vessels [35 36]. Pigmentary abnormalities were defined as areas of increased or decreased pigmentation that did not meet the criteria for geographic atrophy [35 36]. Inter-grader agreement for these two features has previously been reported [35]. For both geographic atrophy and pigmentary abnormalities, label transfer was used between the CFP and (i) their corresponding FAF images and (ii) the CFP-FAF image pairs.

Figure 1 Details of the framework: M3 deep learning convolutional neural network (CNN) for the detection of reticular pseudodrusen from color fundus photographs (CFP) alone, their corresponding fundus autofluorescence (FAF) images alone, or the CFP-FAF image pairs. In the first stage (I. Multi-task learning), the three models (CFP, FAF, and CFP-FAF models) are trained simultaneously. The CFP model takes a CFP as its input, and the FAF model takes the corresponding FAF image as its input. Each image is processed by a CNN backbone, followed by an attention module to capture important image features. (For the CNN backbone, the figure shows a simplified Inception module, but this can be replaced by others, such as ResNet). The important features captured from the CFP and FAF images form the basis of the CFP-FAF model, using cross-modality attention. In the second stage (II. Cascading fine-tuning), the three models are further fine-tuned individually based on the outputs of the first stage. The CFP and FAF models are trained first. The attention modules of the finalized CFP and FAF models are used to extract features for fine-tuning the CFP-FAF model.

*Multi-modal, multi-task, multi-attention deep learning framework*

The proposed deep learning framework is shown in Figure 1, including the multi-modal and multi-task nature of training and the multi-attention mechanism. The framework consists of three deep learning models: the CFP model, the FAF model, and the CFP-FAF model. The CFP model takes CFP images as its input and predicts RPD presence/absence as its output; the same idea applies to the FAF and CFP-FAF models. For the CFP model and FAF model, each has a CNN to extract features from the input image, followed by an attention module to analyze the features that contribute most to decision-making, followed by fully-connected layers, and an output layer, which makes the prediction. The CFP-FAF model has the same structure except that, instead of having its own CNN backbone, it receives the image features from both the CFP and the FAF models. Multi-task training was used to train the deep learning models [37-39]. As shown in Figure 1, this consists of (i) multi-task learning and (ii) cascading task fine-tuning. In multi-task learning, the models are trained jointly, with each model considered as a parallel task, using a shared representation. In cascading task fine-tuning, each model then undergoes additional training separately. The aim of multi-task learning is to learn generalizable and shared representations for all the image scenarios, and the aim of cascading task fine-tuning is to perform additional training suitable for each separate image scenario.

We created three non-M3 deep learning models, one for each image scenario, in order to compare performance between these and the M3 models. Importantly, the structure of the non-M3 CFP model and the non-M3 FAF model represents what is used in existing studies of other medical computer vision tasks to achieve state-of-the-art performance [21 22]. Hence, we created non-M3 models expected to have a high level of performance, in order to set a high standard for the M3 models. For the CFP-only and FAF-only image scenarios, the non-M3 model comprised a CNN backbone, followed by the fully-connected and output layers. To ensure a fair comparison, we used InceptionV3 as the CNN backbone for both the non-M3 and the M3 models [22]. InceptionV3 is a state-of-the-art CNN architecture that is used commonly in medical computer vision applications. For the same reasons, the fully-connected and output layers were exactly the same as those used in the M3 models. For the CFP-FAF image scenario, the non-M3 model used a typical concatenation to combine the CFP and FAF image features from the InceptionV3 CNN backbones. Unlike the M3 models, the three non-M3 models were trained separately and did not use attention mechanisms.

For each of the three image scenarios, we trained an M3 model 10 times using the same training/validation/test split shown in Table 1, to create 10 individual M3 models (i.e. 30 models in total). Similarly, we trained each non-M3 model 10 times (i.e. another 30 models), using the same training/validation/test split. This was to allow a fair comparison between the two model types, including meaningful statistical analysis (as described below). Both the M3 and the non-M3 models shared the same hyperparameters and training procedures to ensure a fair comparison (except that the M3 models had an additional cascading task fine-tuning step, as shown in Figure 1). The InceptionV3 CNN backbones were pre-trained using ImageNet, an image database of over 14 million natural images with corresponding labels, using methods described previously [14]. During the training process, each input image was scaled to 512x512 pixels. The model parameters were updated using the Adam optimizer (learning rate of 0.001) for every minibatch of 16 images. We applied an early stop procedure to avoid overfitting: the training was stopped if the loss on the validation set no longer decreased for 5 epochs. The M3 models completed training within 30 epochs, whereas the non-M3 model completed training within 10 epochs. In addition, image augmentation procedures were used, as follows, in order to increase the dataset size and to strengthen model generalizability: (i) rotation (0-180 degree), (ii) horizontal flip, and (iii) vertical flip. For the cascading task fine-tuning step of the M3 models, the same hyperparameters were used except for a learning rate of 0.0001. The models were implemented using Keras [41] and TensorFlow[42]. All experiments were conducted on a server with 32 Intel Xeon CPUs, using three NVIDIA GeForce GTX 1080 Ti 11Gb GPUs for training and testing, with 512Gb available in RAM memory.

*Evaluation of the deep learning models in comparison with each other*

For the RPD feature, each model was evaluated against the gold standard reading center grades on the full test set of images. For each model, the following metrics were calculated: F1-score, area under receiver operating characteristic (AUROC), sensitivity (also known as recall), specificity, Cohen's kappa, accuracy, and precision. The F1-score (which incorporates sensitivity and precision into a single metric) was the primary performance metric. The AUROC was the secondary performance metric. The performance of the deep learning models was evaluated separately for the three imaging scenarios; for each scenario, the performance of the M3 models was compared with those of the non-M3 models. The Wilcoxon rank sum test was used to compare the F1-scores of the 10 M3 and 10 non-M3 models (separately for each imaging modality). In addition, the differential performance of the models was analyzed by examining the distribution of cases correctly classified by both models, neither model, the non-M3 model only, or the M3 model only. For these analyses, bootstrapping was performed with 50 iterations, with one of the 10 models selected randomly for each iteration. Similar methods were followed for the other two AMD features (geographic atrophy and pigmentary abnormalities).

*Evaluation of the deep learning models in comparison with human ophthalmologists*

For the RPD feature, for each of the three image scenarios, the performance of the deep learning models was compared with the performance of 13 ophthalmologists who manually graded the same images (when viewed on a computer screen at full image resolution). For this comparison, the test set of images was a random subset of the full test set (at the participant level) and comprised 100 CFP, and the 100 corresponding FAF images, from 100 different participants (comprising 68 positive cases and 32 negative cases). The ophthalmologists performed the grading independently of each other, and separately for the two image scenarios (i.e. CFP-alone then FAF-alone). The ophthalmologists comprised three different levels of seniority and specialization in retinal disease: 'attending' level (highest seniority) specializing in retinal disease, attending level not specializing in retinal disease, and 'fellow' level (lowest seniority). Prior to grading, all the ophthalmologists were provided with the same RPD imaging definitions as those used by the reading center graders (i.e. as described above). The performance metrics were calculated, the Wilcoxon rank sum test applied, and ROC curves generated, as above.

*Attention maps*

For the RPD feature, attention maps were generated to investigate the image locations that contributed most to decision-making by the deep learning models. This was done by back-projecting the last convolutional layer of the neural network. The keras-vis package was used to generate the attention maps [43].

*External validation of deep learning models using a secondary dataset not involved in model training*

A secondary and separate dataset was used to perform external validation of the trained deep learning models in the detection of RPD. The secondary dataset was the dataset of images, labels, and accompanying clinical information from a previously published analysis of RPD in the Rotterdam Study. This dataset has been described in detail previously [44]. In this prior study, eyes with and without RPD were selected from the Rotterdam Study, a prospective cohort study investigating risk factors for chronic diseases in the elderly. The study adhered to the tenets in the Declaration of Helsinki and institutional review board approval was obtained.

The dataset comprised 278 eyes of 230 patients aged 65 years and older, selected from the last examination round of the Rotterdam Study and for whom three image modalities were available (CFP, FAF, and NIR) [44]. The positive cases comprised all those eyes in which RPD were detected from CFP (n=72 eyes); RPD presence was confirmed on both FAF and NIR images. The negative cases comprised eyes with soft drusen and no RPD (n=108) and eyes with neither soft drusen nor RPD (i.e. no AMD;

n=98); RPD absence was required on all three image modalities (i.e. CFP, FAF, and NIR). The ground truth labels for RPD presence/absence came from human expert graders locally in the Rotterdam Study. RPD were defined as indistinct, yellowish interlacing networks with a width of 125 to 250 µm on CFP[45]; groups of hypoautofluorescent lesions in regular patterns on FAF [46-48], and groups of hyporeflectant lesions against a mildly hyperreflectant background in regular patterns on NIR images [48].

Each deep learning model was evaluated against the gold standard grades on the full set of images (n=278). As for the primary dataset, the performance of the deep learning models was evaluated separately for the three imaging scenarios. The same performance metrics were used as above.

# RESULTS

## Automated detection of reticular pseudodrusen by multi-modal, multi-task, multi-attention (M3) deep learning

Figure 2. Box plots showing the F1 score results of the M3 and standard (non-M3) deep learning convolutional neural networks for the detection of reticular pseudodrusen from color fundus photographs (CFP) alone, their corresponding fundus autofluorescence (FAF) images alone, or the CFP-FAF image pairs, using the full test set. Each model was trained and tested 10 times (i.e. 60 models in total), using the same training and testing images each time. The horizontal line represents the median F1 score and the boxes represent the first and third quartiles. The whiskers represent quartile 1 - (1.5 x interquartile range) and quartile 3 + (1.5 x interquartile range). The dots represent the individual F1 scores for each model. ****: $P \leqslant 0.0001$; ***: $P \leqslant 0.001$ (Wilcoxon rank-sum test). Note that the Y-axis of the CFP scenario is different.

The results are shown in Figure 2 and Table 2. The F1-scores were substantially higher for the FAF and CFP-FAF scenarios than for the CFP scenario. In all three image scenarios, the F1-score of the M3 model was significantly and substantially higher than that of the non-M3 model. This was particularly noticeable for the clinically important CFP scenario, with an increase of over 20% in the F1-score for the M3 model versus the non-M3 model (60.28 vs. 49.60; p<0.0001). In the FAF scenario, the median F1-scores were 79.30 (IQR 1.41) and 75.18 (IQR 1.94), respectively (p<0.001). In the CFP-FAF scenario, the median F1-scores were 79.67 (IQR 1.19) and 76.62 (IQR 1.71), respectively (p<0.001). The F1-score of the most accurate M3 model, among all runs, was 63.45 for CFP, 79.91 for FAF, and 80.61 for CFP-FAF. The equivalent AUROC values were 84.20, 93.55, and 93.76, respectively. Model calibration analyses were also performed. Figure S1 shows the results, using CFP images as an example. Both the M3 and non-M3 models were moderately well calibrated; the M3 models had a numerically superior Brier score (0.13 vs 0.16).

Figure 3. Differential performance analysis: distribution of test set images correctly classified by both models, neither model, the M3 model only, or the non-M3 model only, for the detection of reticular pseudodrusen from color fundus photographs (CFP) alone, their corresponding fundus autofluorescence (FAF) images alone, or the CFP-FAF image pairs, using the full test set. Each model was trained and tested 10 times (i.e. 60 models in total), using the same training and testing images each time. For each modality, a bootstrapping analysis was performed under 95% confidence interval (randomly selecting one M3 model and one non-M3 model), computing the above distributions, and repeating it for 200 iterations). The mean and standard deviation are shown.

Table 2. Performance results of the M3 and standard (non-M3) deep learning convolutional neural networks for the detection of reticular pseudodrusen from color fundus photographs (CFP) alone, their corresponding fundus autofluorescence (FAF) images alone, or the CFP-FAF image pairs, using the full test set. Each model was trained and tested 10 times (i.e. 60 models in total), using the same training and testing images each time. The median and interquartile range (brackets) are shown for each performance metric.

| | F1-score | Precision | Sensitivity (Recall) | Specificity | AUROC | Kappa | Accuracy |
|---|---|---|---|---|---|---|---|
| **CFP modality** | | | | | | | |
| Standard (non-M3) | 49.60 (11.87) | 61.72 (13.00) | 42.13 (18.21) | 90.40 (7.55) | 77.50 (3.10) | 33.87 (10.00) | 76.74 (1.72) |
| M3 | 60.28 (2.98) | 66.39 (3.11) | 55.28 (9.21) | 89.62 (2.54) | 82.17 (1.05) | 46.49 (2.22) | 79.71 (1.02) |
| **FAF modality** | | | | | | | |
| Standard (non-M3) | 75.18 (1.94) | 86.32 (4.15) | 67.13 (6.84) | 95.94 (1.72) | 91.39 (1.08) | 67.31 (2.63) | 87.76 (1.37) |
| M3 | 79.30 (1.41) | 81.90 (2.91) | 76.72 (3.99) | 93.64 (1.62) | 93.06 (0.49) | 71.71 (2.27) | 88.83 (1.02) |
| **Combined modality** | | | | | | | |
| Standard (non-M3) | 76.62 (1.71) | 81.93 (6.86) | 71.66 (5.39) | 93.85 (2.69) | 91.53 (0.41) | 68.16 (2.11) | 87.97 (0.92) |
| M3 | 79.67 (1.19) | 80.38 (4.18) | 79.42 (3.39) | 92.70 (2.40) | 93.30 (0.46) | 71.90 (1.78) | 88.80 (0.80) |

As observed in Table 2, using the same default cut-off threshold of 0.5, the sensitivity of the M3 models was substantially higher for all three image scenarios, and particularly for the CFP scenario. In addition, the M3 had higher AUROC for all three image scenarios, suggesting that the M3 could better distinguish positive and negative cases. The differential performance of the models was further analyzed by examining the distribution of cases correctly classified by both models, neither model, the M3 model only, or the non-M3 model only, as shown in Figure 3. Analysis of the positive cases demonstrated a relatively high frequency where only the M3 model was correct, particularly for the CFP image scenario (mean 23.7%, SD 9.1%), and a very low frequency of cases where only the non-M3 model was correct (mean 6.1%, SD 4.1%). Similarly, in the FAF scenario, the equivalent figures were 14.2% (SD 6.1%) and 2.1% (SD 1.1%), respectively.

In order to assess whether the multi-modal/multi-task or the multi-attention mechanism contributed most to improved performance of the M3 models, the performance of non-M3 models with one or the other mechanism was examined. The results are shown in Table S1, using CFP images as an example. The F1-score had an absolute increase of 10% (multi-modal/multi-task only) and 6% (multi-attention only), which suggests that both aspects contributed to improved performance (while multi-attention operation also improves model interpretability).

Figure 4. Deep learning attention maps overlaid on representative image examples (color fundus photographs (CFP) alone, fundus autofluorescence (FAF) images alone, or the CFP-FAF image pairs), for the detection of reticular pseudodrusen (RPD) by the M3 model or the non-M3 model: representative examples where the non-M3 model missed RPD presence but the M3 model correctly detected it. For each image, the attention maps demonstrate quantitatively the relative contributions made by each pixel to the detection decision. The heatmap scale for the attention maps is also shown: signal range from -1.00 (purple) to +1.00 (brown). RPD are observed on the FAF images as ribbon-like patterns of round and oval hypoautofluorescent lesions with intervening areas of normal and increased autofluorescence. Areas of RPD clearly apparent to human experts are shown (black arrows), as well as areas of RPD possibly apparent to human experts (dotted black arrows).

Attention maps were generated and superimposed on the fundus images. For each image, these demonstrate quantitatively the relative contributions made by each pixel to the detection decision. Figure 4 shows representative examples where the non-M3 model missed RPD presence but the M3 model correctly detected it. In general, for all three image scenarios, the non-M3 models had only one or very few focal areas of high signal; often, these did not correspond with retinal areas where RPD are typically located. By contrast, the M3 models tended to demonstrate more widespread areas of high signal that corresponded well with retinal areas where RPD are located (e.g. peripheral macula).

## Performance of multi-modal, multi-task, multi-attention (M3) deep learning models versus ophthalmologists in detecting reticular pseudodrusen

For the CFP-alone and FAF-alone image scenarios, the M3 and non-M3 models were used to analyze images from a random subset of the test set. The performance metrics for the detection of RPD were compared to those obtained by each of 13 ophthalmologists.

Figure 5. Receiver operating characteristic (ROC) curves of the M3 and standard (non-M3) deep learning convolutional neural networks for the detection of reticular pseudodrusen from (A) color fundus photographs (CFP) alone or (B) their corresponding fundus autofluorescence (FAF) images alone, using a random subset of the test set (100 CFP and the 100 corresponding FAF images, from 100 different participants). Each model was trained and tested 10 times, using the same training and testing images each time. The mean ROC curve is shown (dotted line), together with its standard deviation (shaded area). The performance of the 13 ophthalmologists on the same test sets is shown by 13 single points. The ophthalmologists comprised three different levels of seniority and specialization in retinal disease: 'attending' level (highest seniority) specializing in retinal disease, attending level not specializing in retinal disease, and 'fellow' level (lowest seniority).

The results are shown in Figure 5 and Table 3. In Figure 5, the performance of the deep learning models is shown by their ROC curves, with the performance of each ophthalmologist shown as a single point. For the CFP scenario, the median F1-scores of the ophthalmologists were 31.14 (IQR 10.43), 35.04 (IQR 5.34), and 40.00 (IQR 9.64), for the attending (retina), attending (non-retina), and fellow levels, respectively. This low level of human performance was expected, since RPD are typically observed very poorly on CFP, even at the gold standard level of reading center experts[9-11]. In comparison, the median F1-score was 64.35 (IQR 6.29) for the M3 models and 49.14 (IQR 24.58) for the non-M3 models. Considering all 13 ophthalmologists together, the F1-scores of the M3 models were approximately 84% higher than those of the ophthalmologists ($p<0.0001$). Indeed, the performance of the M3 models was twice as high as that of the retinal specialists at attending level (the most senior level of ophthalmologists and those specialized in retinal disease).

For the FAF-alone image scenario, the median F1-scores of the ophthalmologists were 81.81 (IQR 3.43), 68.32 (IQR 5.86), and 79.41 (IQR 4.83), for the attending (retina), attending (non-retina), and fellow levels, respectively. In comparison, the median F1-score was 85.25 (IQR 5.24) for the M3 models and 78.51 (IQR 8.51) for the non-M3 models. Considering all 13 ophthalmologists together, the F1-scores of the M3 models were significantly higher than those of the ophthalmologists ($p<0.001$). By contrast, this was not true of the non-M3 models ($p=0.95$). Similarly, the performance of the M3 models was substantially superior to that of all three levels of ophthalmologists considered separately, including the most senior and specialized in retinal disease. Again, this was not true of the non-M3 models.

Table 3. Performance results of the M3 and standard (non-M3) deep learning convolutional neural networks, in comparison with those of 13 ophthalmologists, for the detection of reticular pseudodrusen from color fundus photographs (CFP) alone or their corresponding fundus autofluorescence (FAF) images alone, using a random subset of the test set (100 CFP and the 100 corresponding FAF images, from 100 different participants). Each model was trained and tested 10 times, using the same training and testing images each time. The ophthalmologists comprised three different levels of seniority and specialization in retinal disease: 'attending' level (highest seniority) specializing in retinal disease (4 people), attending level not specializing in retinal disease (4 people), and 'fellow' level (lowest seniority) (5 people). The median and interquartile range (brackets) are shown for each performance metric.

|  | F1-score | Precision | Sensitivity (Recall) | Specificity | AUROC | Kappa | Accuracy |
|---|---|---|---|---|---|---|---|
| **CFP modality** | | | | | | | |
| **Human level** | | | | | | | |
| Fellow | 40.00 (9.64) | 44.44 (10.21) | 37.50 (12.50) | 76.47 (2.94) | - | 11.89 (12.82) | 65.00 (5.00) |
| Attending Other | 35.04 (5.34) | 40.18 (11.68) | 32.81 (10.16) | 80.15 (10.66) | - | 9.01 (11.18) | 63.50 (7.00) |
| Attending Retina | 31.14 (10.43) | 65.91 (39.49) | 21.88 (2.34) | 94.12 (11.76) | - | 18.68 (20.12) | 71.00 (9.75) |
| Overall | 35.00 (9.64) | 44.44 (15.38) | 28.12 (15.62) | 77.94 (16.18) | - | 11.89 (14.24) | 65.00 (8.00) |
| **Model level** | | | | | | | |
| Standard (non-M3) | 49.14 (24.58) | 71.43 (17.13) | 43.75 (27.34) | 91.18 (10.66) | 82.58 (5.09) | 29.28 (20.40) | 73.00 (6.25) |
| M3 | 64.35 (6.29) | 70.19 (6.05) | 57.81 (11.72) | 88.24 (2.21) | 85.66 (2.82) | 48.14 (6.85) | 79.00 (3.50) |
| **FAF modality** | | | | | | | |
| **Human level** | | | | | | | |
| Fellow | 79.41 (4.83) | 71.43 (4.21) | 78.12 (6.25) | 85.29 (4.41) | - | 67.92 (6.98) | 85.00 (5.00) |
| Attending Other | 68.32 (5.86) | 73.05 (6.69) | 68.75 (10.94) | 86.76 (7.72) | - | 54.84 (9.00) | 81.00 (4.25) |
| Attending Retina | 81.81 (3.43) | 91.25 (12.91) | 70.31 (5.47) | 96.32 (5.88) | - | 75.17 (4.51) | 90.00 (1.75) |
| Overall | 79.41 (12.63) | 74.07 (14.78) | 75.00 (12.50) | 88.24 (8.82) | - | 67.92 (18.85) | 85.00 (8.00) |
| **Model level** | | | | | | | |
| Standard (non-M3) | 78.51 (8.51) | 92.67 (7.28) | 65.62 (10.16) | 97.79 (2.57) | 94.18 (2.82) | 71.07 (11.98) | 88.50 (5.00) |
| M3 | 85.25 (5.24) | 91.26 (4.52) | 81.25 (2.34) | 96.32 (1.47) | 95.56 (2.24) | 78.79 (7.59) | 91.00 (3.25) |

*External validation of the multi-modal, multi-task, multi-attention (M3) deep learning models for detecting reticular pseudodrusen: Rotterdam Study*

External validation of the M3 models was performed by testing performance on a secondary and separate dataset of images from the Rotterdam Study. The results are shown in Table 4. The F1-scores of the three M3 models were 78.74 (CFP-alone), 65.63 (FAF-alone), and 79.69 (paired CFP-FAF). The equivalent AUROC values were 96.51, 90.83, and 95.03, respectively. Hence, the performance of the CFP M3 model demonstrated very robust external validation, with performance on the external dataset that was actually substantially higher than on the primary dataset. The F1-score of the FAF M3 model was inferior on the external dataset, and AUROC was modestly inferior. The F1-score of the CFP-FAF M3 model on the external dataset was very similar to that for the primary dataset, and the AUROC was actually superior on the external dataset.

Table 4. External validation of the M3 deep learning convolutional neural network for the detection of reticular pseudodrusen: performance results from color fundus photographs (CFP) alone, their corresponding fundus autofluorescence (FAF) images alone, or the CFP-FAF image pairs, using a test set from the Rotterdam Study. For external validation, we used the model that achieved the highest F1 score on the internal test set.

|           | F1-score | Precision | Sensitivity (Recall) | Specificity | AUROC | Kappa | Accuracy |
|-----------|----------|-----------|----------------------|-------------|-------|-------|----------|
| CFP       | 78.74    | 94.34     | 67.57                | 98.53       | 96.51 | 72.67 | 90.29    |
| FAF       | 65.63    | 77.78     | 56.76                | 94.12       | 90.83 | 55.69 | 84.17    |
| CFP & FAF | 79.69    | 94.44     | 68.92                | 98.53       | 95.03 | 73.80 | 90.65    |

### *Automated detection of geographic atrophy and pigmentary abnormalities by multi-modal, multi-task, multi-attention (M3) deep learning models*

M3 deep learning models were trained to detect two other important features of AMD, geographic atrophy and pigmentary abnormalities. The results are shown in Table S2. For the detection of geographic atrophy, in all three image scenarios, the median F1-scores of the M3 models were numerically higher than those of the non-M3 models. The differences were statistically significant for the CFP-only and FAF-only scenarios ($p<0.001$ and $p<0.01$, respectively). The superiority of the M3 models was particularly evident for the clinically important CFP-only scenario. In the CFP-only scenario, the median F1-score was 83.99 (IQR 1.80) for the M3 model and 80.20 (IQR 1.48) for the non-M3 model. The model with the highest F1-score was the M3 model in the CFP-FAF scenario, at 85.45 (IQR 1.24).

For the detection of pigmentary abnormalities, again, in all three image scenarios, the median F1-scores of the M3 models were numerically higher than those of the non-M3 models. The differences were statistically significant for the FAF-only and CFP-FAF scenarios ($p<0.05$ and $p<0.0001$, respectively). The model with the highest F1-score was the M3 model in the CFP-FAF scenario, at 88.79 (IQR 0.50).

## DISCUSSION

*Clinical importance and implications*

The ability to detect RPD presence accurately but accessibly is clinically important for multiple reasons. RPD are now recognized as a key AMD lesion [6 7]. Their presence is strongly associated with increased risk of progression to late AMD [6]. Identifying these eyes with high likelihood of progression is essential so that clinicians can intervene in a timely way to decrease risk of visual loss. These clinical interventions include prescribing medications (e.g. AREDS2 oral supplements) [49 50], smoking cessation [51], dietary interventions[52], tailored home monitoring [53-55], and tailored reimaging regimens [56]. Importantly, RPD presence is suggested as the critical determinant of the ability of subthreshold nanosecond laser to decrease progression from intermediate to late AMD [57]. However, current attempts to incorporate this key lesion into AMD classification and risk prediction algorithms are hampered. Since RPD grading requires access to both multi-modal imaging and expert graders, ascertainment is limited to the research setting in specialist centers only. Since multi-modal imaging is not typically performed in routine clinical practice, the ability to detect RPD presence from CFP alone represents a valuable step forward in accessibility.

For this clinically important task of detecting RPD (and other common AMD features), we developed and tested a new deep learning approach that benefits from multi-modal, multi-task, and multi-attention operation. This novel M3 approach means that the models can detect RPD accurately in three different scenarios; the approach works irrespective of whether CFP alone, FAF alone, or both imaging modalities are available. Importantly, even when the approach is used in the CFP-alone scenario, the multi-modal and multi-task training mean that the performance benefits from both image types having been present during training. To demonstrate this, we compared with standard non-M3 models where models were trained and tested on exactly the same images as the M3 models. In all three image scenarios, the performance of the M3 models was superior to that of the non-M3 models. This was particularly true for the CFP scenario, the most clinically important task for improving accessibility to RPD grading.

We compared deep learning and human performance, using a large number of ophthalmologists at three different levels of seniority and specialization and from two different institutions. Human performance at detecting RPD from CFP alone was very poor, as expected. Interestingly, human performance on CFP was also relatively independent of seniority/specialization (Table 3). Low performance on CFP by the retinal specialists at attending level was driven particularly by one specialist with very low performance. For detection from CFP or from FAF, the performance of the M3 models was substantially superior to those of the ophthalmologists, including the most senior and specialized group of ophthalmologists.

*Generalizability: external validation and applicability to other important features of age-related macular degeneration*

The M3 models were highly generalizable during external validation. This was shown robustly for all three image scenarios by testing their performance on an independent, well-curated RPD dataset from a different continent. The performance metrics were very high for all three image scenarios. In the case of the CFP scenario, the most important task clinically, performance (AUROC 96.51) was actually higher than during internal testing. That is, despite the fact that the M3 models was trained using AREDS2 images (a dataset with a different population distribution than that of the Rotterdam Study) and had not seen the Rotterdam Study images previously, they still had superior performance to machine learning algorithms previously reported. This high degree of generalizability was likely obtained through the wide breadth of training data used, since the images were obtained from 66 different retinal specialty clinics across the US, comprising a large variety of patients, fundus cameras, and photographers.

In addition, the M3 approach was applicable to important AMD disease features other than RPD, namely geographic atrophy and pigmentary abnormities. Using single-modality, single-task training, we previously demonstrated that deep learning can detect these two features from CFP with similar or slightly superior accuracy to attending level ophthalmologists specializing in retinal disease [14 15]. However, the M3 approach was modestly superior to the non-M3 models for both geographic atrophy and pigmentary abnormalities in all three image scenarios. Again, this improves the accessibility of grading in common clinical scenarios where only CFP is available. Regarding the two previous multi-modal deep learning studies described above [29 30], both comprised training and testing on one disease or disease stage only, with relatively small datasets. More broadly, in the medical image analysis domain, the generalization capability of deep learning models is an open issue: few studies have evaluated the performance of deep learning models in different tasks or with external validation datasets [58 59].

### *Potential advantages of multi-task and multi-attention training*

Multi-task training has important advantages over traditional single-task learning, where each model is trained separately [37]. Single-task training has the disadvantage that the performance of each model is limited by the features present on that particular image modality. Models trained in this way may also be more susceptible to overfitting [40]. By contrast, multi-task training exploits the similarities (shared image features) and differences (task-specific image features) between the features present on the different image modalities. In this way, it usually has improved learning efficiency and accuracy. Essentially, what is learned for each image modality task can assist during training for the other image modality tasks. In this way, it benefits each model by sharing features that are generalizable between the image modalities. We considered that this may be particularly relevant for retinal lesions like RPD, where different imaging modalities (CFP and FAF) highlight very different features relating to the same underlying anatomy.

In addition, many existing multi-modality deep learning models simply concatenate features from each image modality. However, CFP and FAF are very different modalities and they have substantially different features. To address this problem, we employed self-attention [19] and cross-modality attention [20] modules, combined with the multi-task training (Figure 1). For the CFP and FAF models, the self-attention module was used to find the most important features extracted from the CNN backbones. Then, the cross-modality attention module was used to combine the features learnt from the self-attention modules. Importantly, the two self-attention modules (from each of the two image modalities) are shared between all three models.

### *Other strengths, limitations, and further steps required for clinical application*

In addition to those described above, other strengths of this study include the large size and well-characterized nature of the cohort. These data represent one of the largest datasets available where individuals with AMD are followed longitudinally with both CFP and FAF at all study visits. Regarding the ground truth labels, the study also benefits from centralized grading of all images for all relevant AMD features by at least two expert graders at a single reading center with standardized grading definitions [11]. The availability of corresponding CFP and FAF images for all eyes also meant that label transfer between image modalities could be used for training and testing: the ground truth for RPD presence came from the FAF images (but the labels were transferred to the corresponding CFP images); the opposite was true for geographic atrophy and pigmentary abnormalities. Additional strengths include comparison with human performance using a large number of ophthalmologists at three different levels of seniority and specialization in retinal disease.

The limitations include the use of a single imaging modality (FAF) for the ground truth of RPD presence. For RPD detection, NIR imaging may have slightly higher sensitivity than FAF imaging in some studies, though at the expense of lower specificity [60]; NIR imaging may also have low sensitivity in detecting ribbon-type RPD [61]. Similarly, optical coherence tomography (OCT) imaging is reported to have slightly higher sensitivity and specificity than FAF imaging for RPD detection. However, this may depend on the OCT device, and standard macular OCT scans may miss some cases where the RPD are more peripheral [6]. Ideally, multi-modal imaging may be used for detecting RPD. However, adding a second imaging modality (e.g. FAF with NIR, or FAF with OCT) can increase either the sensitivity or the specificity, but not both [6]. One potential limitation is the use of multiple images from individual eyes (from sequential annual study visits), from approximately half of the eyes. Although this was done in order to increase the data available for training, it can increase the chance of overfitting to the internal dataset. However, this concern is much less relevant, given that the models appeared highly robust during external validation testing.

Further steps need to be taken for the findings to be applied in the clinic. As well as the successful external validation described here, we are planning additional external validation testing using data from other datasets. Prospective validation would also be useful, comparing the performance of deep learning and ophthalmologists in the detection of RPD in a prospective clinical trial setting. This approach was the basis for the first FDA approval of an autonomous artificial intelligence system in clinical care, for the detection of diabetic retinopathy [62].

**CONCLUSIONS**

In this work, we presented M3, a multi-modal, multi-task, multi-attention deep learning framework for the detection and detailed characterization of AMD. In all three image scenarios (CFP only, FAF only, and CFP-FAF), M3 performance was significantly superior to that of existing deep learning methods and of human experts. This was particularly true in the clinically important CFP scenario, where its performance was twice as high as that of the retinal specialists. The M3 approach was also highly generalizable: during external validation on an independent dataset from a different continent, RPD detection from CFP alone was highly accurate. Generalizability was also shown by adaptation to detecting two other important AMD features.

Overall, we believe that this M3 approach demonstrates the potential for automated but accurate ascertainment of the full spectrum of AMD features from CFP alone. This is extremely valuable for improved AMD classification and risk prediction. Importantly, operation from CFP alone makes it accessible far beyond the small number of specialist centers in the developed world with access to multi-modal imaging and expert graders. Planned future work consists of additional external validation and prospective assessment in a clinical trial setting. This would take a step towards clinically applicable artificial intelligence systems.

**CONFLICT OF INTEREST STATEMENT**

Q.C., T.D.K., Y.P., E.A., W.T.W., E.Y.C., and Z.L. are co-inventors on patent applications based on artificial intelligence methods applied to age-related macular degeneration ('Methods and Systems for Predicting Rates of Progression of Age-Related Macular Degeneration'; E-057-2020-0-US-01; E-058-2020-0-US-01).

The other authors declare that there are no competing interests.

**AUTHOR CONTRIBUTIONS**

Conception and design: Q.C., T.D.K., E.Y.C., and Z.L. Analysis and interpretation: Q.C., T.D.K., A.A., Y.P., E.A., C.C.K., D.T.L., M.H.C., C.A.C., H.E.W., T.M., C.C.K., W.T.W., Y.Y.Z., E.Y.C., and Z.L. Data collection: Q.C., T.D.K, and A.D. Drafting the work: Q.C., T.D.K., E.Y.C., and Z.L.

**APPENDIX**

The other members of the AREDS2 Deep Learning Research Group include:

Malena Daich, MD, David Dao, MD, Lucas Groves, MD, Chris Hwang, MD, Henry Lin, MD, Alisa Thavikulwat, MD, and Anton Vlasov, MD.

**DATA AVAILABILITY**

The AREDS2 dataset containing the data analyzed and generated during the current study has been deposited in dbGAP, accession number phs002015.v1.p1. Following completion of the current Institutional Review Board review of the dataset (i.e., in approximately three months), the data will be freely available from dbGAP. In the intervening period, the dataset is available from the corresponding author on reasonable request.

**ACKNOWLEDGEMENTS**

The work was supported by the intramural program funds and contracts from the National Center for Biotechnology Information/National Library of Medicine/National Institutes of Health, the National Eye Institute/National Institutes of Health, Department of Health and Human Services, Bethesda, Maryland (Contract HHS-N-260-2005-00007-C; ADB contract NO1-EY-5-0007; Grant No K99LM013001). Funds were generously contributed to these contracts by the following National Institutes of Health: Office of Dietary Supplements, National Center for Complementary and Alternative Medicine; National Institute on Aging; National Heart, Lung, and Blood Institute; and National Institute of Neurological Disorders and Stroke.